\documentclass{article}

\usepackage[preprint,nonatbib]{neurips_2022_ml4ad}

\usepackage[utf8]{inputenc} 
\usepackage[T1]{fontenc}    
\usepackage{url}            
\usepackage{booktabs}       
\usepackage{amsfonts}       
\usepackage{nicefrac}       
\usepackage{microtype}      
\usepackage{xcolor}         

\usepackage{epsfig}
\usepackage{graphicx}
\usepackage{threeparttable}

\title{Detection of Active Emergency Vehicles \\using Per-Frame CNNs and Output Smoothing}

\author{%
  Meng Fan, Craig Bidstrup, Zhaoen Su, Jason Owens, Gary Yang, Nemanja Djuric \\
  Aurora Innovation, Inc.\\
  \texttt{\{meng.fan, cbidstrup, zhaoen, jlowens, garyyang, ndjuric\}@aurora.tech} \\
}

\begin{document}

\bibliographystyle{ieeetr}

\maketitle

\begin{abstract}
While inferring common actor states (such as position or velocity) is an important and well-explored task of the perception system aboard a self-driving vehicle (SDV), it may not always provide sufficient information to the SDV.
This is especially true in the case of active emergency vehicles (EVs), where light-based signals also need to be captured to provide a full context.
We consider this problem and propose a sequential methodology for the detection of active EVs, using an off-the-shelf CNN model operating at a frame level and a downstream smoother that accounts for the temporal aspect of flashing EV lights.
We also explore model improvements through data augmentation and training with additional hard samples. 
\end{abstract}

\section{Introduction}

Self-driving vehicles (SDVs) rely on an onboard perception system to understand their surroundings, which includes detection of all traffic actors in the vicinity as well as their tracking through time.
The output of this system commonly includes actor type (such as vehicle, pedestrian, bicyclist, etc.), as well as the state of the actor (such as bounding box size, position, velocity, acceleration, etc.), which is a well-studied problem in the literature \cite{fernandes2021point, chen2017multi}.
However, these outputs may not capture all relevant attributes for some traffic actors that the SDV may interact with.
This is the case when it comes to the detection of active emergency vehicles (EVs), where the state of the emergency lights provides an important additional piece of information to the SDV.

Vision-based detection of EVs has been a topic of several recent studies.
Most researchers proposed to first use an object detector implemented as a convolutional neural network (CNN) that is then run on an input camera image to detect vehicle actors, followed by an EV-specific classifier operating on cropped images of detected actors. 
In particular, the authors of \cite{roy2019emergency} proposed to use YOLO \cite{redmon2016you} to detect vehicles and crop images around them, and then apply a VGG-16-based EV classifier trained on individual vehicle images. 
In \cite{baghel2020analysis} the authors introduced a similar two-stage approach with an object detector and a CNN classifier, which is trained and tested on a small custom data set, while \cite{tran2021audio} used a single-stage model by training an object detector YOLOv4 \cite{bochkovskiy2020yolov4} directly on a data set of two object classes: EV and non-EV. 
In \cite{kherraki2022deep, haque2022emergency} the authors explored several existing CNN architectures (such as DenseNet~\cite{huang2017densely}) for EV classifications using image crops of individual vehicles as inputs, trained and evaluated on smaller data.
In addition, the authors of \cite{haque2022emergency} showed the benefits of using pre-trained weights obtained by training on ImageNet data \cite{deng2009imagenet}, followed by a fine-tuning step used to specialize the models for EV-specific tasks.
The authors in \cite{razalli2020emergency} extracted color values from the top of an image crop (where the EV siren lights are expected to be found) provided by a vehicle detector and applied a support vector machine for classification.

In these earlier works the data came either from non-EV-specific public sources (such as MS COCO \cite{lin2014microsoft}) or from EV-specific YouTube channels \cite{youtube_1,youtube_2,youtube_3} and CCTV footage. 
The statistics in such data sets are far from what we find in the real world though. 
For example, the EV-to-non-EV ratio is unrealistically high in data sets collected from EV-specific YouTube channels, and in general the data will not be able to cover the numerous non-EV types and appearances commonly observed on the roads. 
Moreover, such data are usually biased towards close-range EVs that are easier to classify, while for the SDV use-case one would like to detect EVs at the first sight at long ranges, which presents a more challenging task.
Besides, the previous approaches did not consider the EV activeness (i.e., whether or not the EV lights are flashing).
In this work we adopt the two-step approach discussed above while addressing several limitations of the prior studies. 
We collect on-road data from our self-driving fleet covering rich scenes involving EVs, including day and night, short and long ranges, highway and urban, and various other traffic conditions.
We focus on detecting active EVs that are more important for SDVs to respond to.
In addition, we apply approaches such as data augmentation, hard-sample mining, and output smoothing, which help further boost the performance.

An alternative approach worth mentioning is audio-based detection \cite{tran2021audio, tran2020acoustic, cantarini2022few, asif2022large}, which does not require an unobstructed line of sight to detect active EVs. 
However, on highways and in general at higher ego-speeds acoustic approaches are significantly affected by ambient noise (such as wind). We do not consider such methods in our current work and instead focus on image-based approaches.

\section{Data set and labeling}
\label{section:labeling}

We collected on-road data logs using an autonomous driving fleet equipped with lidar and camera sensors. 
For all vehicles we perform 4D labeling (i.e., a sequence of 3D bounding boxes) and 2D labeling (i.e., a 2D bounding box within a camera image). 
We associate 4D and 2D labels and generate a collage video of each vehicle using the camera images. We then use it to label the actor into one of 4 categories: police car, ambulance, fire truck, and non-EV. 
If the vehicle is labeled as an EV it will trigger the second-stage labeling for activeness and bulb state. 
In particular, at each time frame we label the vehicle as active if the beacon light is flashing, and inactive otherwise. For active frames, we label it as "bulb on" if the light is illuminated and "bulb off" otherwise. Note that the activeness labeling relies on the local temporal context, while the bulb state labeling only relies on a single frame. 

\begin{table}
  \caption{Data set summary}
  \label{table_data}
  \centering
  \begin{tabular}{ll}
    \toprule
    \cmidrule(r){1-2}
    Category     & Distribution \\
    \midrule
    Time of logs &  Day: 81.1\%, Night: 18.9\%   \\ 
    Vehicle type &  EV: 3.4\%, non-EV: 96.6\%   \\ 
    EV type & police vehicle: 80.0\%, fire vehicle: 13.4\%, ambulance: 6.6\% \\
    EV activeness & active: 90.0\%, inactive: 10.0\% \\
    Bulb state of active EVs & bulb-on: 91.8\%, bulb-off: 8.2\% \\
    \bottomrule
  \end{tabular}
\end{table}

We collected logs containing a total of around $12{,}000$ unique vehicles, each with a maximum duration of 25 seconds.
As seen in the data summary shown in Table \ref{table_data}, we observed a large fraction of non-EVs, which is more realistic than previous studies.
We also find that the majority of EVs are active. 
Active EVs usually have multiple beacon bulbs flashing, and thus when looking at each single frame there is a small chance ($8.2\%$) to observe all bulbs being off. 
This motivates us to design a single-frame detector for active EV actors, which also has faster inference than a multi-frame detector.

\section{Methodology}

\subsection{The EV detection system}

The EV detection system is composed of three phases: pre-processing, inference, and post-processing. 
During pre-processing, a tracker system produces tracks of the surrounding vehicles that are used as input. 
We project the 3D coordinates of the corners of each vehicle track’s bounding box into the camera image and compute the centroid. 
For each centroid within the image bounds, we use the smallest axis-aligned square that encapsulates the projected corners to crop the image and resize it to $224 \times 224$ pixels. 
We ignore tracks where the centroid is out of the field of view of the forward camera, as well as those with a projected width smaller than a certain threshold (set to $18$ pixels in this study). 
We then batch the valid image patches and forward them to the inference module. 
During inference, we run the EV classifier model (discussed in detail in the next section) on the image patches and output probabilities indicating whether a certain vehicle is an active EV or not. 
Then, in order to capture the temporal component of active flashing lights, we implemented a downstream smoother in the post-processing phase that infers the final overall state. 
The smoother is a cyclic buffer that keeps a ledger of the last $25$ valid EV outputs for each vehicle track’s history and outputs a smoothed result.
We require that there are at least $6$ detected frames of the actor and more than $T = 50\%$ of per-frame active EV outputs in the buffer to mark the vehicle as being an active EV. 
This helps suppress transient positive outputs and thus mitigate false positives (FPs), discussed further in Section \ref{section:results}.
The EV detection system can handle hundreds of actors (processed as a batch) in real-time, with an average latency of less than $10$ milliseconds.

\subsection{The EV classifier model}
\label{section:model}

The EV classifier is a CNN using ResNet-18 (layers 1 to 4) \cite{he2016deep} as a backbone to extract image features, followed by a global average pooling layer and a fully-connected layer with a single output channel that is passed through sigmoid to output probability. 
The ResNet backbone uses the weights pre-trained on the ImageNet data \cite{deng2009imagenet} and is not frozen during training. 
We set the label as positive when a vehicle is an EV with a “bulb on” state at the time of image capture, and as negative otherwise.
We use focal loss \cite{lin2017focal} due to a large data imbalance.
The Adam optimizer with $0$ weight decay and an initial learning rate of $10^{-4}$ is used. 
We apply a customized optimizer wrapper that decreases the learning rate when there is no loss improvement for a certain number of iterations, and stops training when the learning rate drops to a given low value. 
The collected logs are split randomly into train and test sets with a $3:1$ ratio.
The model is implemented in PyTorch \cite{paszke2019pytorch} and trained on a single GPU.

\begin{table}[t]
  \caption{Per-frame classifier results (A: data augmentation; M: mined data from data engine)}
  \label{table_model}
  \centering
  \begin{threeparttable}
  \begin{tabular}{lcc}
    \toprule
    \cmidrule(r){1-2}
    Model     & \% change of max-F1 & \% change of precision at $0.8$ recall\\
    \midrule
    Baseline & 0  & 0 \\
    + A & 3.11  & 5.3  \\
    + A + M  & 5.73 & 10.4 \\
    \bottomrule
  \end{tabular}
  \begin{tablenotes}[para,flushleft]
  \end{tablenotes}
  \end{threeparttable}
\end{table}

\subsection{Techniques for data improvement}

The considered problem is non-trivial due to the imbalance between EVs and non-EVs and challenging image artifacts, especially at long ranges. 
We applied several techniques to address these issues.

\subsubsection{Data augmentation}
\label{section:data_aug}

Data augmentation is often applied when training an image classifier by modifying raw images using operations such as cropping, flipping, rotation, resizing, and others. 
We augment only the rare active-EV class and focus on approaches that do not introduce unrealistic deviations and additional bias between the classes.
Our solution updates the vehicle track bounding box as follows.
The track state contains $x$, $y$, and $z$ coordinates of the bounding box center, as well as its length, width, and height. 
To each of these states we fit a normal distribution on the training set.
Then, given an input track, we sample from this distribution to generate new bounding boxes, ensuring the augmented data is likely to be seen in the real world. 
When creating the final train data set we use a sampling ratio of $2\times$ on positive examples, and also downsample negative examples by $5\times$.

\subsubsection{Data engine: Mine, label, and retrain}

As EVs are very rare compared to other traffic actors, we explored additional sourcing methods to find relevant data in the collected data logs. 
Another practical problem is that the proposed model may produce FPs, and we would like to collect such hard samples to help train a better classifier. 
To address these two problems, we developed a data engine to continuously improve the learned model.
In particular, we mine the data in the newly collected logs to find the events with active-EV detections by the EV classifier, regardless of whether or not they were true or false positives. 
Then, we label such events and add them to the data set, before retraining the EV classifier model.

\section{Results}
\label{section:results}

\begin{figure}[t]
  \includegraphics[scale=0.395]{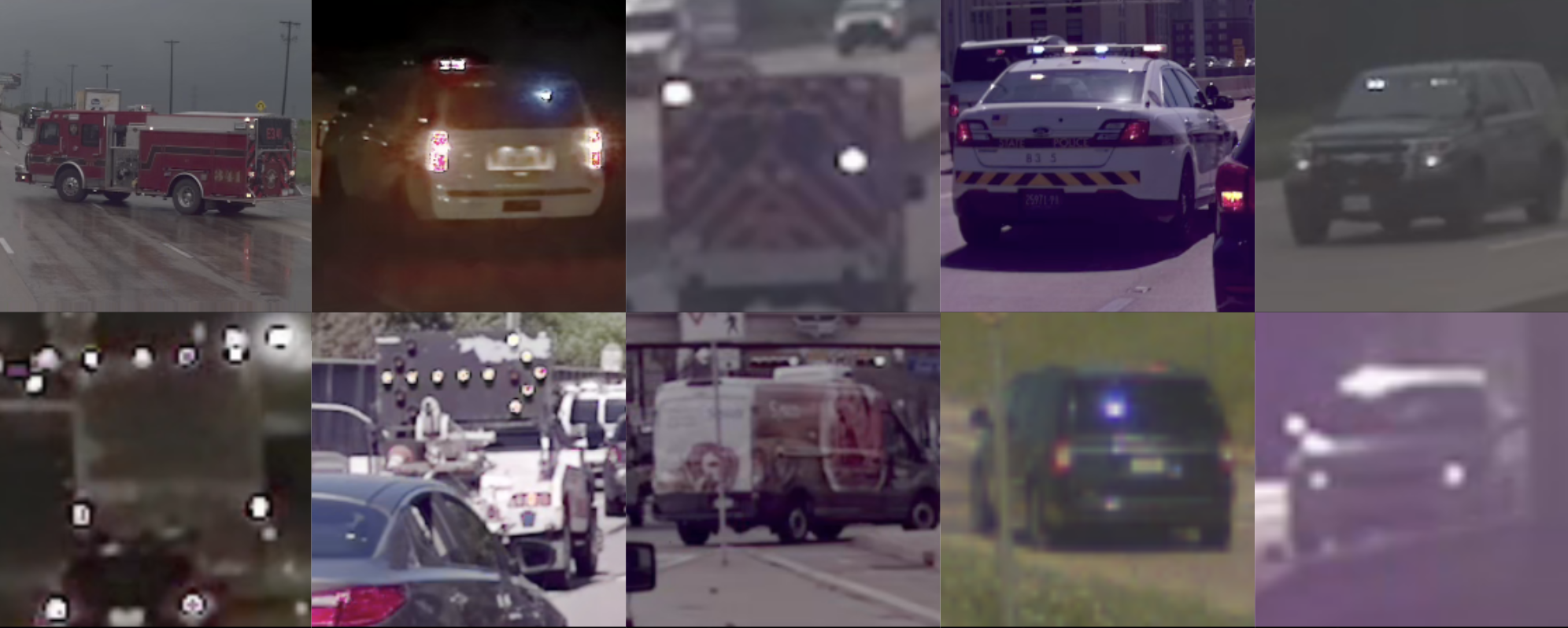}
  \centering
  \caption{Example detections (top row: true positives, bottom row: difficult true negatives)}
  \label{fig:examples}
\end{figure}

\begin{table}[t]
  \caption{Per-actor results of the output smoother}
  \label{table_smoother}
  \centering
  \begin{threeparttable}
  \begin{tabular}{cccc}
    \toprule
    \cmidrule(r){1-2}
    Smoother threshold $T$ & \% change of precision & \% change of recall & \% change of F1 score \\
    \midrule
    $0\%$  & 0 & 0 & 0 \\
    $30\%$  & 10.30 & -0.68 & 5.07 \\
    $50\%$  & 9.87 &  -0.51 & 4.87 \\
    $70\%$  & 6.80 & -20.0 & -7.02 \\
    \bottomrule
  \end{tabular}
  \begin{tablenotes}[para,flushleft]
  \end{tablenotes}
  \end{threeparttable}
\end{table}

We performed hyper-parameter tuning to train a strong initial model, which is used as a baseline which other experiments are compared to. 
The experimental results are shown in Table~\ref{table_model}. 
By using data augmentation discussed in Section \ref{section:data_aug}, we see $3.11\%$ boost in the max-F1 score compared to the baseline. 
If we also add more training data collected through hard-sample mining (nearly doubling the number of vehicles) and retrain the model, we see an improvement of $5.73\%$ in max-F1. 
Moreover, this method improves the precision by more than $10\%$ at a fixed $0.8$ recall. 
More importantly, the data engine helps to continuously iterate and improve the model by adding useful data as it is collected.

In Figure~\ref{fig:examples} we show examples of true positives (TPs) and difficult true negatives (TNs) of the improved "+ A + M" model from Table \ref{table_model}. 
The TPs that are shown in the first row cover different types of EVs, various weather and light conditions, as well as multiple viewing angles. 
The difficult negative cases (shown in the second row) were false positives of the baseline model, but then successfully predicted as TNs by the improved model. 
It is interesting to note that such difficult cases were usually caused by some other light sources, including brake lights from vehicles queued ahead, flashing lights of construction vehicles, or sun reflections. 

Beyond the learned EV model, we also evaluated the performance of the downstream smoother. 
Using a different setup of the smoother we show the corresponding changes in precision, recall, and F1 score in Table~\ref{table_smoother}, where we tune the threshold $T$ (i.e., the minimum fraction of positive outputs in the buffer required to output a final positive state). 
As can be seen, when the threshold is $30\%$ and $50\%$, we reach a higher F1 score and precision with very small regression in recall, which shows the effectiveness of the smoother in mitigating FPs. 
However, when the threshold is set to a value that is too high (such as $70\%$), the system starts to generate more FNs which hurts the overall performance. 

\section{Conclusion}
We considered the problem of detecting active emergency vehicles in the context of self-driving vehicles. 
To address this task we proposed to use a frame-level EV detector whose outputs are fed to an output smoother, which captures the temporal dimension of such actors.
Finally, we evaluated the method on large-scale, real-world data, and made further improvements by data augmentation and by training with additional hard samples mined using a data engine.

{
\small
\bibliography{ev}
}
\end{document}